\documentclass{article}

\PassOptionsToPackage{numbers, compress}{natbib}

    \usepackage[preprint]{neurips_2022}

\usepackage[utf8]{inputenc} 
\usepackage[T1]{fontenc}    
\usepackage{hyperref}       
\usepackage{url}            
\usepackage{booktabs}       
\usepackage{amsfonts}       
\usepackage{nicefrac}       
\usepackage{microtype}      
\usepackage{xcolor}         
\usepackage{graphicx}
\usepackage{bm}
\usepackage{amsmath}
\usepackage{multirow}

\title{Swing Distillation: A Privacy-Preserving Knowledge Distillation Framework}

\author{
Junzhuo Li$^1$\footnotemark[2], Xinwei Wu$^2$\footnotemark[2], Weilong Dong$^2$\footnotemark[2], Shuangzhi Wu$^3$, Chao Bian$^3$, Deyi Xiong$^{2,1}$\footnotemark[1]\\
$^1$School of New Media and Communication, Tianjin University, Tianjin, China\\
$^2$College of Intelligence and Computing, Tianjin University, Tianjin, China\\
$^3$ByteDance Lark AI,  Beijing, China\\
\texttt{\{jzli, wuxw2010, willowd, dyxiong\}@tju.edu.cn}\\
        \texttt{wufurui@bytedance.com, chaobian@outlook.com}
}

\begin{document}

\maketitle

\begin{abstract}
  Knowledge distillation (KD) has been widely used for model compression and knowledge transfer.
Typically, a big teacher model trained on sufficient data transfers knowledge to a small student model.
However, despite the success of KD, little effort has been made to study whether KD leaks the training data of the teacher model. 
In this paper, we experimentally reveal that KD suffers from the risk of privacy leakage. To alleviate this issue, we propose a novel knowledge distillation method, swing distillation, which can effectively protect the private information of the teacher model from flowing to the student model.
In our framework, the temperature coefficient is dynamically and adaptively adjusted according to the degree of private information contained in the data, rather than a predefined constant hyperparameter. 
It assigns different temperatures to tokens according to the likelihood that a token in a position contains private information.
In addition, we inject noise into soft targets provided to the student model, in order to avoid unshielded knowledge transfer.
Experiments on multiple datasets and tasks demonstrate that the proposed swing distillation can significantly reduce (by over 80\% in terms of canary exposure) the risk of privacy leakage in comparison to KD with competitive or better performance. Furthermore, swing distillation is robust against the increasing privacy budget.
\end{abstract}
\renewcommand{\thefootnote}{\fnsymbol{footnote}}
\footnotetext[2]{Contribution during internship at ByteDance Lark AI.}
\footnotetext[1]{Corresponding author.}
\renewcommand{\thefootnote}{\arabic{footnote}}
\section{Introduction}

Data is usually privacy-sensitive and not always publically available.
High-resource parties usually own a huge amount of labeled data for training models, which, however, is not the case for low-resource institutions or parties.
Intuitively, there are two ways to bridge the gap between high- and low-resource parties with respect to model training: data sharing and model sharing. The former directly shares data across parties at the high risk of privacy leakage. The latter usually fine-tunes the shared model obtained from a high-resource party on its own small labeled data, which prevents direct data exposure. 

However, recent studies \cite{carlini2021extracting,xu2021beyond,mireshghallah2022memorization} have found that attacking shared neural models can lead to the exposure of training data. This suggests that directly sharing models across parties is still at the risk of privacy leakage. 
Our research question is hence how we can allow low-resource parties to obtain high-quality models on the premise of data security.

Knowledge distillation is originally proposed to solve the problem of model compression \cite{hinton2015distilling}. But it has also been widely used for model training in data-constrained scenarios as aforementioned. The key idea is to transfer “knowledge” from a teacher model to a student model.
If the teacher model is well trained on sufficient labeled data, it can well teach the student model via knowledge transfer, although the data of the student model is very limited.

Therefore, using KD for cross-party learning seems to be natural and straightforward  as the training data of the teacher model is not directly shared with the student model. 
However, there has been no research on whether KD is able to protect the private information of the teacher model from being transferred to the student model.

\begin{figure}[t] 
    \centering
    \small
    \includegraphics[width=0.48\textwidth]{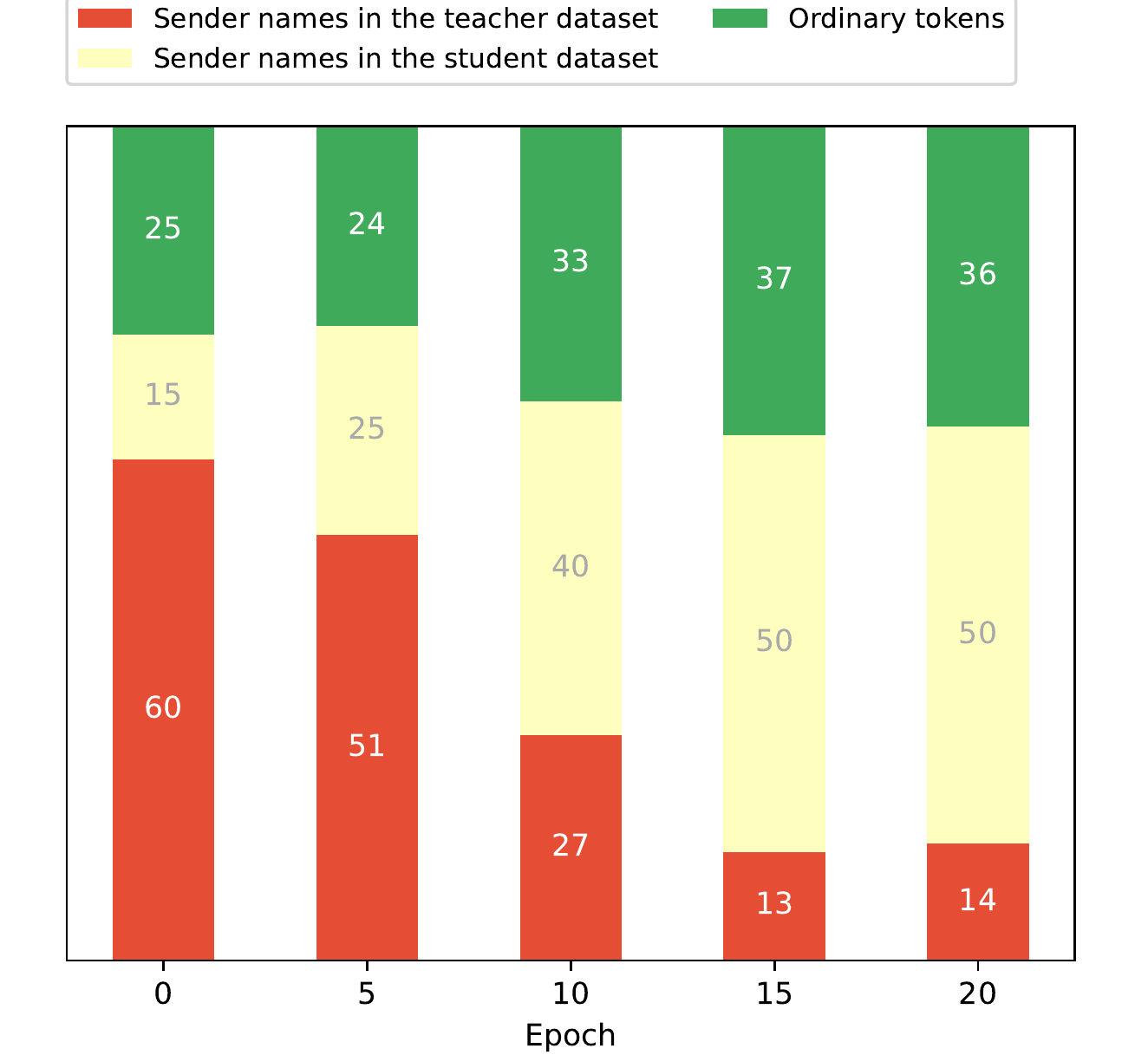}
    \caption{The distribution of different types of tokens (sender names in the teacher dataset vs. those in the student dataset vs. ordinary tokens) in the top-100 tokens with the highest probabilities predicted by the student model when the model is attacked at different training stages. The numbers of sender names occurring in the teacher dataset and student dataset are 93 and 50, respectively.}
    \label{fig:sender_name_leakage}
\end{figure}

In order to investigate this question, we have conducted distillation experiments on the AESLC dataset \cite{zhang-tetreault-2019-email}, which is a dataset for email subject line generation. We divide the dataset into two parts, one for the teacher model and the other for the student model. There is no overlap on the names of email senders in these two parts. After conducting knowledge distillation (see Section~\ref{sec:privacy_leakage_KD} for more details), we find that even if the names of email senders in the teacher dataset are not present in the student dataset at all, the student model after distillation still has a high probability to output these names, as shown in Figure~\ref{fig:sender_name_leakage}. This clearly suggests that KD suffers from privacy leakage from the teacher model to the student model. 

To mitigate this issue, we propose a new KD method, {\bf Swing Distillation}, in which the temperature coefficient is no longer a fixed value, but dynamically adjusted according to data privacy.
We force the student model to selectively learn the knowledge of the teacher model by varying the temperature of the adjusted distillation.
In addition, we perform a special privacy protection operation by injecting noise into the soft targets transferred to the student model, which can further protect the training data of the teacher model from being exposed.
Experiments on multiple datasets and tasks show that swing distillation can significantly reduce the risk of leakage of private information while ensuring the distillation effect.
This provides a safe way for knowledge transfer across parties.

\paragraph{Contributions} Our main contributions are as follows:
\begin{itemize}
    
    \item We experimentally reveal that knowledge distillation suffers the risk of privacy leakage of the teacher model, though KD can improve the performance of student models.
    
    \item We propose a new distillation method, Swing Distillation (SD) that distills knowledge of the teacher model to the student model while protecting the privacy of the training data  of the teacher model. The proposed SD includes two essential components: dynamic temperature and soft target protection.

    \item Experiment results show that our method achieves performance competitive to or even better than that of KD on a variety of datasets/tasks, and significantly reduces the exposure of the private information in the teacher model training data by over 80\%.
\end{itemize}

\section{Preliminary}

\paragraph*{Knowledge Distillation}
\citet{hinton2015distilling} introduced soft targets (i.e., probabilities over class labels with a hyperparameter $T$) and propose the knowledge distillation for model compression:

\begin{equation}\label{eq:pi}
    P_i(\bm{z}_i, T) = \frac{\exp(\bm{z}_i/T)}{\sum_{q=0}^k\exp(\bm{z}_q/T)},
\end{equation}
where $k$ is the number of target classes, $T$ is the temperature coefficient, which is used to control the softening degree of the output probability. Specifically, the distillation loss and student loss can be computed as:
\begin{equation}
\begin{aligned}
    & \mathcal{L}_\text{KD} = -\sum^N_{j=0}\sum_{i=0}^k P_i(\bm{z}_i^{(j)}, T) \log (P_i(\bm{v}_i^{(j)}, T)),\\
    & \mathcal{L}_\text{S} = -\sum^N_{j=0}\sum_{i=0}^k \bm{y}_i^{(j)}\log(P_i(\bm{v}_i^{(j)}, 1)),
\end{aligned}
\end{equation}
where $\bm{z}$ and $\bm{v}$ are the logits of the teacher and student model, respectively, $\bm{y}$ is the ground-truth label, and $N$ is the total number of samples. The total loss of knowledge distillation is the linear interpolation of the above two losses: 
\begin{equation}
    \mathcal{L} = \lambda \mathcal{L}_\text{KD} + (1-\lambda)\mathcal{L}_\text{S},
\end{equation}
where $\lambda$ is a hyperparameter. The value of $\lambda$ is usually fixed after being tuned on a development set.

Usually, $T$ is set to 1 during testing while a higher $T$ is used during training. When $T=1$ during testing, the soft targets of different classes vary greatly, so during testing it can better distinguish the correct class from the incorrect classes. During training, the differences between soft targets with a higher $T$ are smaller than those when $T=1$, and the model will have more emphasis on the incorrect classes with smaller probabilities. In this way, the student model learns from both correct and incorrect classes.

\paragraph*{Privacy Leakage in KD}\label{sec:privacy_leakage_KD}

\citet{carlini2021extracting} have verified that neural language models can memorize training data, and output instances from the original training data when given inductive prompts. In KD, the teacher model transfers soft targets as knowledge to the student model. If the soft targets contain sensitive information, it is likely for the teacher model to transfer the privacy of its own training data to the student model, resulting in privacy leakage.

We conducted experiments on the AESLC dataset \cite{zhang-tetreault-2019-email} to investigate the privacy leakage phenomenon in knowledge distillation. AESLC is a dataset constructed from the Enron Email Data \cite{bryan-2004enron}. We inserted "This email was written by [X]." at the end of each email, where [X] is a placeholder for the sender's name. We then divided the dataset into two subsets of different sizes according to the name of email senders, ensuring that each email sender name will only appear in one subset.
We used the large subset to train a teacher model, the small subset to train a student model. The trained teacher model performed knowledge distillation to the student model on the small dataset.

We used "This email was written by" as a prompt to check if the student model would output the sender name inserted into the teacher dataset in the placeholder.
Figure~\ref{fig:sender_name_leakage} presents the distribution of the top-100 tokens predicted by the student model when it is given this prompt.
It can be seen that the student model has a certain probability of generating sender names from the teacher dataset when it encounters an attack. As these names are not present in the student dataset, only occurring in the teacher dataset, these results strongly suggest that KD is at risk of leaking private information to the student model. 
\begin{figure*}[t]
    \centering
    \small
    \includegraphics[width=0.98\textwidth]{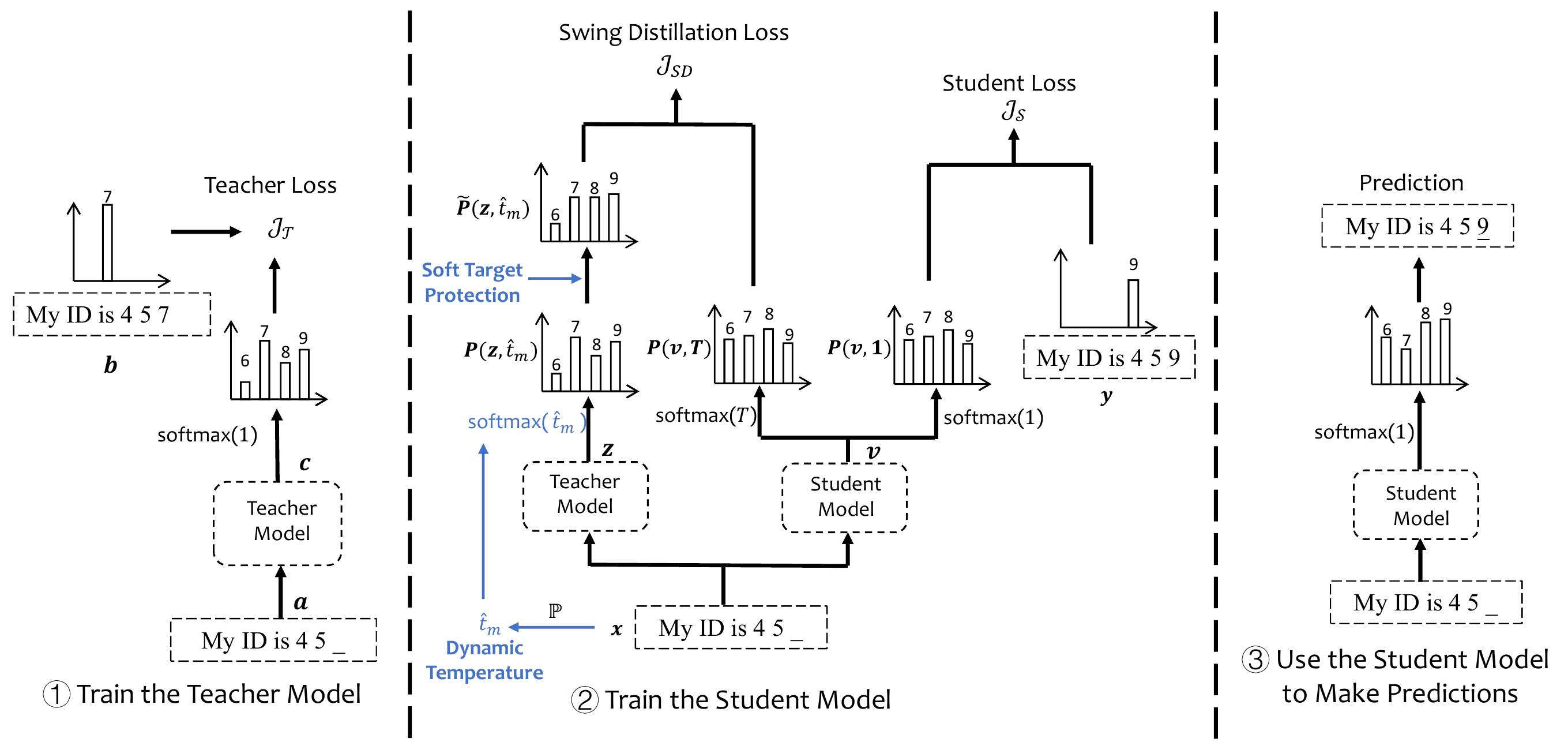}
    \caption{The diagram of Swing Distillation. The two proposed strategies are in blue.
}
    \label{fig:overview}
\end{figure*}
\section{Swing Distillation}

In order to avoid privacy leakage in KD, we propose a privacy-preserving knowledge distillation framework, \textbf{S}wing \textbf{D}istillation (SD). Figure~\ref{fig:overview} illustrates the diagram of SD. SD mainly introduces two strategies over KD to prevent the leakage of privacy from the teacher model to the student model: dynamic temperature and soft target protection. The two strategies protect different types of private information. 

\subsection{Teacher Model Training}
The teacher model is trained on the teacher dataset $\mathbb{T}$, where each data instance consists of an input $\bm{a}$ and ground-truth label $\bm{b}$.
$\bm{a}$ is input into the teacher model to get the output $\bm{c}$, i.e., $\bm{c} = \mathcal{T}(\bm{a})$.

The loss function of the teacher model can be defined as: 
\begin{equation}
    \mathcal{J}_\mathcal{T} = -\sum^{|\mathbb{T}|}_{j=0}\sum_{i=0}^k \bm{b}_i^{(j)}\log(P_i(\bm{c}_i^{(j)}, 1)),
\end{equation}
where $k$ is the number of classes. In a generation task, $k$ is the size of vocabulary.

After the teacher model $\mathcal{T}$ is trained, its parameters are frozen to ensure that the teacher model will not be changed during the distillation phase.

\subsection{Dynamic Temperature}
In KD \cite{hinton2015distilling}, the temperature is used to control the sharpness of the distribution of soft targets. A lower temperature sharpens the distribution of soft targets and hence enables the model to pay more attention to classes with maximal logits while a higher temperature makes the distribution flat and hence increases the difficulty of learning \cite{li2022curriculum,chandrasegaran2022revisiting,liu2022meta}. This inspires us to use a higher temperature for distilling soft targets with private information while a lower temperature for non-sensitive soft targets. 

We hence want to use adaptive temperature coefficients to control the transferrence of soft targets, so that 
the teacher model can "selectively" transfer knowledge to the student model.

The occurrence of private information is often accompanied by specific keywords, such as "\textit{name}", "\textit{password}" and so on. These keywords can be treated as clues for identifying private information, which are referred to as privacy clue words.
We hence construct a privacy clue word dictionary $\mathbb{P}$ to determine what information is not expected to be transferred by the teacher model to the student model (More details on $\mathbb{P}$ are provided in Appendix~\ref{app:privacy_clue_word}).

For a given sequence $\bm{x}=[x_1, x_2, ..., x_n]$, our hypthosis is that if $x_i$ is a privacy clue word, that is, $x_i\in\mathbb{P}$, the closer a token $x_m$ to $x_i$, the more the token contains privacy information.
Therefore, during distillation, the temperature $\hat{t}_m$ corresponding to $x_m$ is estimated:
\begin{equation}
    \hat{t}_m = T + \alpha\frac{n}{|i-m|},
\end{equation}
where $T, \alpha\in\mathbb{R}$ are the temperature coefficient and privacy protection coefficient respectively. The smaller the distance between $x_m$ and $x_i$ (i.e., the higher the likelihood that $x_m$ contains private information), the higher the temperature is (so as to prevent the transfer of privacy).
In this way, we have a set of temperature coefficients in a vector $\bm{\hat{t}}\in\mathbb{R}^n$, rather than a constant $T$.

If there are multiple privacy clue words in the input $\bm{x}$, we choose the closest privacy clue word to the token in question to calculate the temperature.

During distillation, each input $\bm{x}$ from the student dataset $\mathbb{S}$ is fed into the teacher model to obtain the corresponding soft targets. If the token to be predicted is $x_m$, Eq. (\ref{eq:pi}) is reformulated as fowllows:
\begin{equation}
    P_i(\bm{z}_i, \hat{t}_m) = \frac{\exp(\bm{z}_i/\hat{t}_m)}{\sum_{q=0}^k\exp(\bm{z}_q/\hat{t}_m))},
\end{equation}
where 
$\bm{z}$ is the output of the teacher model $\mathcal{T}$, that is, $\bm{z} = \mathcal{T}(\bm{x})$.

\subsection{Soft Target Protection}

Not all private information is associated with a privacy clue word. It is difficult to protect such privacy information by only relying on the dynamic temperature strategy.
Therefore, we further propose a soft target protection strategy to enhance privacy protection in the distillation process.

An intuitive idea is to add Laplacian noise ($\bm{r}\sim\text{Lap}(1/\epsilon)$) to soft targets to make it satisfy the differential privacy condition \cite{Dwork2006DifferentialP} for all data entries $\bm{d}, \bm{d}'\in\mathbb{D} $ and all outputs $\bm{o}\in\mathbb{O}$:
\begin{equation}
    \text{P}[\mathcal{M}(d)=\bm{o}] \leq \exp(\epsilon)\text{P}[\mathcal{M}(d')=\bm{o}],
\end{equation}
where $\mathcal{M}:\mathbb{D}\rightarrow\mathbb{O}$ is a randomised algorithm mapping a data entry in $\mathbb{D}$ to $\mathbb{O}$, and $\epsilon$ is the privacy budget.

When the privacy budget $\epsilon$ is small, the availability of soft targets will be correspondingly reduced, and the teacher model may transfer wrong knowledge to the student model at this time. Moreover, in the task of multi-label classification (the generation task can be regarded as $|\mathbb{V}|$-label classification), the probability distribution of soft targets is extremely unbalanced. In other words, the predicted probabilities mostly distribute over the top few classes. Therefore, it is unnecessary to add noise to the probability of each class.

Therefore, we inject Laplacian noise into the top-K classes with the highest probabilities in $P(\bm{z}, \hat{t}_m)$ to get $\tilde{P}(\bm{z}, \hat{t}_m)$:
\begin{equation}
    \tilde{P}_i(\bm{z}_i, \hat{t}_m) = P(\bm{z}_i, \hat{t}_m) + \bm{r}_i,\\
\end{equation}
where $\bm{r} \in \mathbb{R}^K$ follows the Laplace distribution, and $P(\bm{z}_i, \hat{t}_m)$ denotes the probabilities of class from the top-K classes. The probabilities of the other classes remain unchanged. After this step, we can get the protected soft targets $\tilde{P}(\bm{z}, \hat{t}_m)$.

\subsection{Training and Inference of the Student Model}

During distillation, an input $\bm{x}$ in $\mathbb{S}$ is fed into the teacher model $\mathcal{T}$ and the student model $\mathcal{S}$ to obtain $\bm{z}$ and $\bm{v}$ respectively:
\begin{equation}
    \begin{aligned}
        \bm{z} = \mathcal{T}(\bm{x}),\quad \bm{v} = \mathcal{S}(\bm{x}).
    \end{aligned}
\end{equation}

Following the practice of KD, the softmax with a constant $T$ can be used directly since the data in $\mathbb{S}$ is not private information for the student model itself. The loss function of SD can be defined as:
\begin{equation}
    \mathcal{J}_\text{SD} = -\sum^{|\mathbb{S}|}_{j=0}\sum_{i=0}^k \tilde{P}_i(\bm{z}_i^{(j)}, \hat{t}_m^{(j)}) \log (P_i(\bm{v}_i^{(j)}, T)).
\end{equation}

Similar to KD, the overall training objective is the linear interpolation of the SD loss with the distillation from the teacher model and the cross-entropy loss of the student model learning by itself: 
\begin{equation}
    \begin{aligned}
        & \mathcal{J} = \lambda \mathcal{J}_\text{SD} + (1-\lambda) \mathcal{J}_\mathcal{S}, \\
        & \mathcal{J}_\mathcal{S} = -\sum^{|\mathbb{S}|}_{j=0}\sum_{i=0}^k \bm{y}_i^{(j)}\log(P_i(\bm{v}_i^{(j)}, 1)),
    \end{aligned}
\end{equation}
where $\bm{y}$ is the ground-truth label of $\bm{x}$, and $P_i$ is consistent with the description in Eq. (\ref{eq:pi}).

For inference, predictions can be made locally on the student model. This is only required to set the temperature coefficient $T$ to 1 for the softmax function for prediction in the trained student model.

\begin{table*}[t]
    \centering
    \resizebox{\textwidth}{!}{
    \begin{tabular}{c|c|ccc|ccc}
    \hline
    \multirow{2}*{Datasets} & \multirow{2}*{Models} &\multicolumn{3}{c|}{Text Summarization} & \multicolumn{3}{c}{Canary Attack}\\
    \cline{3-8}
    & & Rouge-1 $\uparrow$&  Rouge-2 $\uparrow$ & Rouge-L $\uparrow$ & Rank $\uparrow$ & Exposure $\downarrow$  & $\Delta \downarrow$ \\
    \hline
    \multirow{5}*{CNN/DailyMail} & 
    Teacher Model & 43.94 & 21.40 & 40.82 & 1 & 19.93 & -\\
    & Student Model& 41.33 & 19.68 & 38.26 & 732127 & \textbf{0.45}  & -\\
    
    & KD & 43.56 & 20.45 & 39.84 & 113426 & 3.14 & +2.65\\
    & SD & 43.43 & 20.81 & 39.92 & 687569 & \underline{0.54} & +0.09 \\
    & Fine-tuning & 43.98 & 21.37 & 40.89 & 1 & 19.93 & +19.48\\
    \hline
    \multirow{5}*{\textsc{BigPatent-e}} & 
    Teacher Model &  44.56 & 16.61 & 38.30 & 1 & 19.93 & -\\
    & Student Model&  40.80 & 14.13 & 34.88 & 701263 &  \textbf{0.51} & - \\
    
    & KD & 41.53 & 14.33 & 35.43 & 112578 & 3.15 & +2.64\\
    & SD & 41.96 & 14.61 & 35.85 & 681498 & \underline{0.55} & +0.04\\
    & Fine-tuning & 44.73 & 16.69 & 38.41 & 1 & 19.93 & +19.42\\
    \hline
    \multirow{2}*{Dataets} & \multirow{2}*{Models} & \multicolumn{3}{c|}{Question Generation} & \multicolumn{3}{c}{Canary Attack} \\\cline{3-8}
    &&Rouge-L $\uparrow$&BLEU-4 $\uparrow$&METEOR  $\uparrow$& Rank  $\uparrow$& Exposure  $\downarrow$&$\Delta \downarrow$ \\\hline
    \multirow{5}*{SQuAD 1.1}& Teacher Model & 48.13 & 20.56 & 25.17 & 1 &19.93 & -\\
    & Student Model &46.39 & 18.45 & 23.61 & 710095 & \textbf{0.49} & -\\
    & KD & 47.85 & 19.62 & 24.80 & 143676 & 2.80 & +2.31\\
    & SD & 47.79 & 19.59 & 24.72 & 673219 & \underline{0.57} & +0.08\\
    & Fine-tuning & 48.72 & 21.17 & 25.36 & 1 & 19.93 & +19.44\\\hline
    \multirow{5}*{MSQG}& Teacher Model & 37.91 & 8.45 & 23.62 & 1 & 19.93 & - \\
    & Student Model & 35.14 & 6.43 & 20.43 & 692345 & \underline{0.53} &-\\
    & KD & 36.75 & 7.84 & 22.51 & 126726 & 2.98 & +2.45 \\
    & SD & 36.69 & 7.79& 22.43 & 706609 & \textbf{0.50} & -0.03\\
    & Fine-tuning & 38.23 & 8.72 & 23.97 & 1 & 19.93 & +19.40\\\hline
    \multirow{2}*{Dataets} & \multirow{2}*{Models} & \multicolumn{3}{c|}{Question Answering} & \multicolumn{3}{c}{Canary Attack} \\\cline{3-8}
    &&\multicolumn{3}{c|}{F1  $\uparrow$} & Rank  $\uparrow$& Exposure  $\downarrow$& $\Delta \downarrow$\\\hline
    \multirow{5}*{CoQA}& Teacher Model & \multicolumn{3}{c|}{66.41} &1 &19.93&-\\
    & Student Model & \multicolumn{3}{c|}{54.83}& 756643 & \textbf{0.40} & -\\
    & KD & \multicolumn{3}{c|}{59.35} & 149008 & 2.75 & +2.35\\
    & SD & \multicolumn{3}{c|}{58.81}& 707236 & \underline{0.50} & +0.10\\
    & Fine-tuning & \multicolumn{3}{c|}{67.27} & 1 & 19.93 & +19.53\\\hline

    \end{tabular}}
    \caption{Main results on the CNN/DailyMail, \textsc{BigPatent-e}, SQuAD 1.1, MSQG, and CoQA dataset. \textbf{Bold} and \underline{underlined} results indicate the lowest and second lowest canary exposure, respectively. $\uparrow$: the higher the better. $\downarrow$: the lower the better. The last column shows the absolute increments of the canary exposure of different models compared with the student model. }
    \label{tab:main_result}
\end{table*}

\section{Experiments}
We carried out experiments and in-depth analyses on different datasets and tasks (i.e., text summarization, question generation and question answering) to validate the effectiveness of the proposed Swing Distillation.
\subsection{Datasets, Tasks and their Evaluation Metrics}
\paragraph*{Text Summarization}
We uesd two datasets with different domains: CNN/DailyMail and \textsc{BigPatent}.
The CNN/DailyMail Dataset \cite{hermann2015teaching} is an English dataset containing over 300K unique news articles written by journalists at CNN and the Daily Mail. 
 The \textsc{BigPatent} dataset \cite{sharma-etal-2019-bigpatent} consists of 1.3 million US patent documents and human written abstracts, which are divided into 9 different categories. Our experiments were mainly carried out on the data of the E (Fixed Constructions) category. 
 We used ROUGE-1/2/L \cite{lin-2004-rouge} as metrics to evaluate all models on this task .

\paragraph*{Question Generation}

For this task, we used SQuAD 1.1 and MSQA. 
The SQuAD 1.1 \cite{rajpurkar-etal-2016-squad} dataset contains over 100K crowdsourced questions with corresponding answer spans extracted from 536 Wikipedia articles. Since the original test set of the SQuAD 1.1 is not publicly available, we follow \citet{du-etal-2017-learning} and \citet{zhao-etal-2018-paragraph} to construct a test set with examples from the original training set and development set. Once these examples were randomly selected into test set, they were removed from the training/development set.
The MSQG dataset \cite{liu-etal-2021-glge} consists of 220K articles from real-world search engines.
Each passage contains a highlight span and a related query. We treat the queries as questions in this dataset.
ROUGE-L, BLEU-4 \cite{papineni-etal-2002-bleu}, and METEOR \cite{banerjee-lavie-2005-meteor} were used as the metrics for this task.

\paragraph*{Question Answering}
We used CoQA \cite{reddy-etal-2019-coqa} in the QA task, which contains 127K questions with answers, obtained from 8K conversations about text passages from seven domains. The input for this task is a sequence of conversation history along with a given question and a given passage, and the target output is a freeform answer text. F1-Score \cite{rajpurkar-etal-2016-squad} was used as the metric to evaluate this task.

For each dataset, we split the training set into a teacher dataset $\mathbb{T}$ and a student dataset $\mathbb{S}$ at a ratio of 19:1. The data statistics of the teacher/student datasets are given in Appendix~\ref{app:experiment} in detail. We train models on different datasets and finally evaluate the performance of the model on the test sets.

\subsection{Baselines}
We compared SD with KD and Fine-tuning in terms of both task performance and privacy-preserving capability. We used BART \cite{lewis-etal-2020-bart}, specifically, \texttt{facebook/bart-base}\footnote{\url{ https://huggingface.co/facebook/bart-base}}, as the backbone model for all methods.
The student model was only trained on $\mathbb{S}$ while the teacher model was trained on $\mathbb{T}$.
The fine-tuning method fine-tuned the teacher model on $\mathbb{S}$, while both KD and SD used the output of the teacher model as soft targets for distillation on $\mathbb{S}$.
Furthermore, we compared methods for protecting soft targets using Laplacian noise (more details can be found in Section~\ref{sec:compare_exp}).
As we focus on the privacy preservation in the knowledge distillation procedure and scenario, we did not compare with other privacy-preserving technologies (e.g., DP-SGD \cite{abadi2016deep}, federated learning \cite{mcmahan2017communication}) that deal with problems and scenarios significantly different from ours.

\subsection{Evaluating Privacy-Preserving Capability via Canary Exposure}
Following previous work \cite{carlini2019secret}, we evaluated the capability of privacy preserving by inserting canary tokens/sequences into the training data. Specifically, we insert special sequences (referred to as canaries) into the training dataset. We train a model on the data with canaries and calculate the exposure of the inserted canaries to measure if the model memorizes these canaries and outout them.

Given a canary $c$, a model with parameters $\bm{\theta}$, and the randomness space $\mathcal{R}$, the
exposure $e_{\bm{\theta}}$ of canary $c$ can be calculated as :

\begin{equation}
    e_{\bm{\theta}} = \log_2|\mathcal{R}|-\log_2 \text{Rank}_{\bm{\theta}}(c).
\end{equation}

We inserted "\textit{My ID is 4 6 7 8 2 3. }" into $\mathbb{T}$ as the canary.
For text summarization, question generation, and question answering tasks, we inserted the canary into summaries, questions, and answers, respectively. This is because these fields are what the trained models are to generate and we can easily detect the inserted canary once it is included in the generated outputs.

\subsection{Main Results}

Table ~\ref{tab:main_result} presents our main results, including both the task performance and the canary exposure.
The results show that SD performs very competitively to, or even outperforms, KD on the three tasks. Importantly, SD significantly reduces canary exposure by over 80\% compared to KD, suggesting that SD is able to significantly improve the privacy protection performance of the model without having a negative impact on the task performance.

Since the canary is not inserted into $\mathbb{S}$, the canary exposure of the student model can be treated as the oracle result. The exposure of the \texttt{facebook/bart-base} model that has not been fine-tuned is 0.45 (very close to the exposure of the student model), illustrating that this is the best exposure result as the two models do not see the canary at all. Our experimental results show that SD can reduce the canary exposure to this oracle exposure. This demonstrates that SD is strongly effective in privacy protection.

From these results, we observe that the canary exposure of KD is much higher than those of the student model and SD, which further verifies that knowledge distillation suffers from privacy leakage.

We notice that the exposure of the teacher model on all datasets is 19.93, which is also the maximum exposure. This is because the teacher model memorizes the inserted canary and output the canary when given a canary-sensitive prompt.
Although directly fine-tuning the teacher model on $\mathbb{S}$ can lead to better task performance, its exposure is 19.93, which is the same as that of the teacher model that is not fine-tuned. This suggests that direct fine-tuning is at a high risk of privacy leakage  (as the fine-tuned model still memorizes the inserted canary), which compromises the improvements on the task performance gained by fine-tuning.

We also observe that SD achieves stable improvements over the student model on task performance and significant gains over KD on the canary exposure across different tasks and datasets. This demonstrates the robustness and applicability of the proposed SD on multiple tasks. 

We visualize temperatures in SD vs. KD in Appendix~\ref{app:temp_vis}.

\begin{table}[t]
    \centering
    \resizebox{0.48\textwidth}{!}{
    \begin{tabular}{c|c|ccc}
        \hline
        Datasets & Metrics & SD & w/o DT & w/o STP \\\hline
        \multirow{5}*{\shortstack{CNN/\\DailyMail}}&R-1 & 43.43 & 43.35 & 43.61 \\
        &R-2 & 20.81 & 20.15 & 20.89 \\
        &R-L & 39.92 & 39.86 & 39.96 \\
        &Exp. & 0.54 & 2.56 & 1.04 \\
        & $\Delta$ & - & +374.1\% & + 92.6\%\\
        \hline
        \multirow{5}*{\shortstack{SQuAD\\1.1}}&R-L & 47.79 & 47.65 & 47.83 \\
        &B-4 & 19.59 & 19.51 & 19.85 \\
        & MTR & 24.72 & 24.69 & 24.92\\
        &Exp. & 0.57 & 2.75 & 1.28 \\
        & $\Delta$ & - & +382.5\% & +124.6\%\\\hline
    \end{tabular}}
    \caption{Ablation study on the CNN/DailyMail and SQuAD 1.1 datasets. The last row of each dataset shows the absolute increments of the canary exposure of different models compared with SD. R-L: ROUGE-L. B-4: BLEU-4. MTR: METEOR. Exp.: Canary Exposure.}
    \label{tab:ablation_study}
\end{table}

\subsection{Ablation Study}

We conducted ablation experiments to examine the effectiveness of the proposed Dynamic Temperature (DT) and soft target protection (STP) strategy. "w/o DT" denotes that the dynamic temperature strategy is not used in SD while "w/o STP" refers to SD without soft target protecton.
Table~\ref{tab:ablation_study} displays the results of ablation experiments on CNN/DailyMail and SQuAD 1.1. Ablation results on all datasets are shown in Appendix~\ref{app:experiment}.

It can be observed that removing the two strategies from SD has a very small impact on the task performance (or even with a better task performance). By contrast, it has a very significant impact on privacy protection in terms of the canary exposure. The absence of the dynamic temperature strategy results in huge canary exposure increments of 350+\% over SD while the exclusion of the soft target protection strategy increases the canary exposure by over 100\% on average. This suggests that the two strategies are important for privacy protection in SD and the dynamic temperature strategy contributes almost 3 times as much as the soft target protection strategy to the privacy preservation.

\begin{table}[t]
    \centering
    \begin{tabular}{lccccc}
    \hline
         & R-1 & R-2 & R-L & Exp. \\\hline
        KD \\ 
        \quad +Laplace & 42.89 & 20.25 & 38.84 & 2.29\\
        \quad +STP & 43.47 & 20.51 & 39.78 & 2.34\\\hline
        DT & \\
        \quad +Laplace & 42.74 & 20.14 & 38.65 & 0.61 \\
        \quad+STP (SD) & 43.43 & 20.81 & 39.92 & 0.54\\\hline
    \end{tabular}
    \caption{The performance of the student model on the CNN/DailyMail dataset with different protection strategies. R-L: Rouge-L. Exp.: Canary Exposure.}
    \label{tab:noise_compare}
\end{table}

\subsection{Injecting Noise of Different Distributions into KD and SD}\label{sec:compare_exp}

We further conducted experiments to compare the effect of injecting noise of different distributions (e.g., Laplacian Distribution) into KD and SD.  Table~\ref{tab:noise_compare} reports the results. The difference between KD and DT (dynamic temperature) is that the former uses a static temperature while the latter takes a dynamic temperature. The difference between "+Laplace" and our STP is that the former inject noise into all soft targets while STP only top-K soft targets ($K$ was set to 3 in all experiments). All these methods use the same privacy budget $\epsilon=1$, which indicates a relatively tight privacy guarantee.

Although Laplacian noise injection can reduce the exposure of the student model, they significantly degrade the performance of the student model on the corresponding task. Especially compared with our proposed STP, the negative impact on the task performance is more pronounced.

This may be due to the small privacy budget, and that "+Laplace" inject noise into all soft targets and might change the distribution, resulting in undesirable knowledge transfer to the student model.

In KD, all the noise injection strategies can effectively reduce the exposure of the model. However, the canary exposure is still higher than that of the student model. This again demonstrates the necessity and effectiveness of the dynamic temperature strategy.

\section{Related Work}

\paragraph*{Knowledge Distillation}

Knowledge distillation has been widely used for model compression and enhancement. Both transfer the knowledge of the teacher model to the student model. The difference is that in the model compression the teacher model guides the training of the student model on the same labeled dataset to obtain a small yet efficient model \cite{lu2017knowledge,sanh2019distilbert,tsai-etal-2019-small,ding2019compression,fu2020ultrafast,liu-etal-2020-fastbert,jiao-etal-2020-tinybert}. 

Model enhancement focuses on using other resources (e.g., unlabeled or cross-modal data) or knowledge distillation optimization strategies (e.g., mutual learning and self-learning) to improve the performance of a student model \cite{markov2016robust,joy2017generalized,abraham2017transfer,shen2018feature,hu-etal-2018-attention,feng-etal-2019-learning,chen-etal-2020-distilling,wu-etal-2020-single}.
There have been studies \cite{jafari-etal-2021-annealing,li2022curriculum,chandrasegaran2022revisiting,liu2022meta} exploring dynamic temperature in KD. Significantly different from ours, their purposes are to improve the performance of the model, instead of protecting privacy in KD.

\paragraph*{Privacy Attack and Protection in Natural Language Processing}
Previous studies have empirically found that deep neural models are at risk of privacy leakages \cite{chang2018privacy,carlini2019secret,pan2020privacy}.

In Natural Language Processing (NLP), three privacy attack methods have been studied.
The first is the extraction attack tailored for the memorization phenomenon of large language models (memorizing training data), which can extract privacy-sensitive texts of the training data from the generative language models \cite{carlini2021extracting}. The canary attack \cite{carlini2019secret} used in our experiments belongs to this category.
The second attack method is the membership inference attack, which predicts a specific attribute or fragment of a sample based on prior information \cite{jagannatha2021membership}. 
The third is an attack based on gradient reduction of training data segments \cite{dimitrov2022lamp}.

In order to deal with privacy concern in NLP models, a variety of methods have been proposed to protect privacy-sensitive information from being leaked. 
These approaches can be roughly divided into three groups according to the stage where they are applied: method used in the data processing stage, in the pre-training and/or fine-tuning stage, and in the post-processing stage.
In the data processing stage, protection is mainly carried out by modifying the input texts, for example, replacing sensitive information in the inputs by named entity recognition \cite{oak-etal-2016-generating,liu2017identification,dernoncourt2017identification,obeid2019impact,eder-etal-2020-code}, anonymizing sensitive information \cite{Arvind2008robust,sanchez-etal-2018-automatic,Mosallanezhad-etal-2019-deep,garcia-pablos-etal-2020-sensitive}, or adding noise or random replacement into the inputs \cite{fernandes2019generalised,melamud-shivade-2019-towards,Qu2021NaturalLU,oluwaseyi-etal-2020-privacy}.
Privacy protection in the pre-training \& fine-tuning stage mainly uses gradient optimization strategies based on differential privacy \cite{coavoux-etal-2018-privacy,melamud-shivade-2019-towards,Li2021LargeLM,xu-etal-2021-utilitarian,qu2021privacy,dupuy2022efficient}. 
The basic idea is to fuse noise into the gradients of each batch of data, thereby reducing the difference between gradients and avoiding the memorization of training data.
The methods used in the post-processing stage are mainly to let the trained model forget specific data or change specific parameters so as to achieve the purpose of protecting the hidden private information in the model \cite{bourtoule2019machine,varun2021advances,Neel2020DescenttoDeleteGM}.

Security protection based on knowledge distillation is to use knowledge distillation to protect the security and privacy of neural models \cite{wang2019privatemodel,vongkulbhisal2019unifying}. 
Direct use of data will inevitably violate privacy. 
Similar to federated learning \cite{mcmahan2017communication}, KD is able to avoid private data being shared across parties.

Particularly, \citet{wang2019privatemodel} transfer the features learned from the private data of the teacher model to the student model through a public dataset. \citet{vongkulbhisal2019unifying} separately train multiple classifiers through knowledge distillation as a unified classifier. The significant difference of our work from these works is that they attempt to train a safe model on private data while our goal is to prevent the privacy in the model trained on private data from being leaked to a student model.

\section{Conclusion}

In this paper, we have empiricaly verifed that KD is at the risk of leaking privacy in the training data of the teacher model to student model. To address this issue, we have presented a new distillation framework, Swing Distillation, where the temperature coefficient is dynamically and adaptively adjusted according to the privacy of each instance. In addition, to avoid privacy leakage through soft targets, we further propose a soft target protection strategy via noise injection. 
Experiments on multiple datasets and tasks demonstrate that (1) SD is capable of significantly reducing the canary exposure of the student model while maintaining the task performance competitive to or even better than KD and (2) both strategies are effective in privacy preservation.

\clearpage
\bibliographystyle{acl}
\bibliography{anthology,custom}

\clearpage

\appendix

\begin{table}[t]
    \centering
    \resizebox{0.48\textwidth}{!}{
    \begin{tabular}{c|c|ccc}
        \hline
        Datasets & Metrics & SD & w/o DT & w/o SLP \\\hline
        \multirow{5}*{\textsc{BigPatent-e}}&R-1 & 41.96 & 41.87 & 42.03 \\
        &R-2 & 14.61 & 14.57 & 14.78 \\
        &R-L & 35.85 & 35.79 & 35.92 \\
        &Exp. & 0.55 & 2.64 & 1.24 \\
        & $\Delta$ & - & +380.0\% & + 125.5\%\\
        \hline
        \multirow{5}*{MSQG}&R-L & 36.69 & 36.58 & 36.72 \\
        &B-4 & 7.79 & 7.61 & 7.81 \\
        & MTR & 22.43 & 22.35 & 22.47 \\
        &Exp. & 0.50 & 2.37 & 1.19 \\
        & $\Delta$ & - & +374.0\% & +138.0\%\\\hline
        \multirow{3}*{CoQA}&F1 & 58.81 & 58.79 & 59.22 \\
        &Exp. & 0.50 & 2.19 & 1.12 \\
        & $\Delta$ & - & +338.0\% & +124.0\%\\\hline
    \end{tabular}}
    \caption{Ablation study on \textsc{BigPatent-e}, MSQG, and CoQA. The last row of each dataset shows the absolute increments of the canary exposure of different models compared with SD. R-L: ROUGE-L. B-4: BLEU-4. MTR: METEOR. Exp.: Canary Exposure.}
    \label{tab:ablation_study_2}
\end{table}

\begin{table*}[t]
    \centering
    \resizebox{\textwidth}{!}{
    \begin{tabular}{cccccccc}
    \hline
        Datasets &  \#Train & \#Dev & \#Test & \#Input & \#Output & Input & Output \\\hline
        CNN/DailyMail & 287, 113 & 13,368 & 11, 490 & 822.3 & 57.9 & article & summary\\
        \textsc{BigPatent-e} & 34, 443 & 1, 914 & 1, 914 & 3572.8 & 116.5 & description & abstract \\
        SQuAD 1.1 & 75, 722 & 10, 570 & 11, 877 & 149.4 & 11.5 & answer/passage & question\\
        MSQG & 198, 058 & 11, 008 & 11, 022 & 45.9 & 5.9 &  highlight/passage & question\\
        CoQA & 108, 647 & 3, 935 & 4, 048 & 354.4 & 2.6 & history/passage & answer\\\hline
    \end{tabular}}
    \caption{Dataset statistics. \#Train/Dev/Test: the number of examples in training/development/test set. \#Input/Output: the average number of tokens in the input/output.}
    \label{tab:statistic}
\end{table*}

\section{Additional Experiments}\label{app:experiment}
We also conducted ablation experiments on the three datasets of \textsc{BigPatent-e}, MSQG, and CoQA, and the results are shown in Table~\ref{tab:ablation_study_2}.

Data statistics of the 5 used datasets are shown in Table~\ref{tab:statistic}.

\section{Temperature Visualization in SD vs. KD}\label{app:temp_vis}

\begin{figure}[t]
    \centering
    \small
    \includegraphics[width=0.48\textwidth]{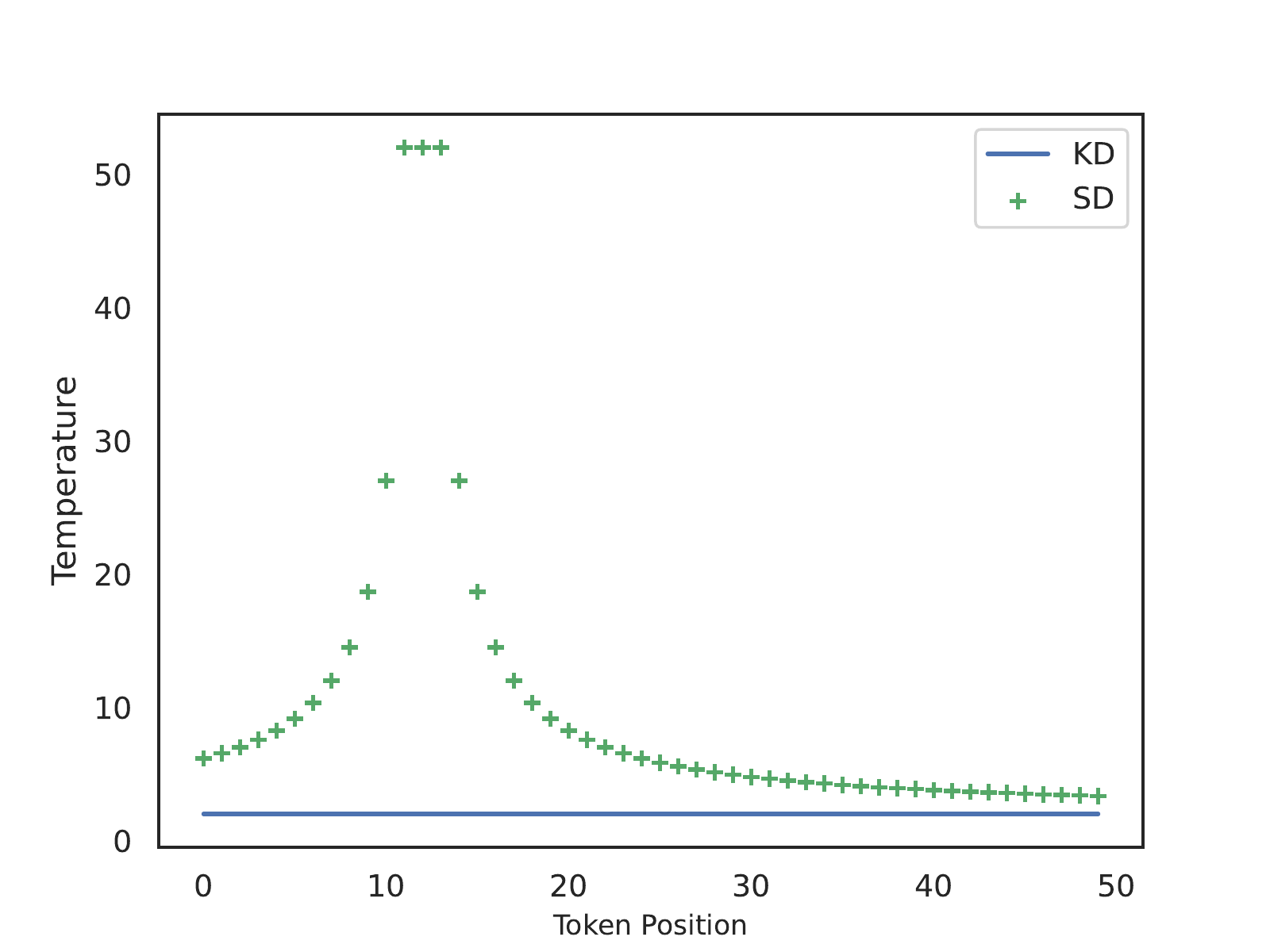}
    \caption{The curve of temperature of Swing Distillation and Knowledge distillation.
}
    \label{fig:temperature}
\end{figure}

We visualize the dynamic temperatures in SD vs. KD (T=2) for an example in Figure~\ref{fig:temperature}. This is an email with 50 tokens, and the 12th token is a privacy clue word.

It can be seen that the temperatures of SD change accordingly, instead of being constant for all tokens like KD. 
Under the dynamic temperature strategy, the temperatures near the privacy clue word are very high. Since $\hat{t}_m\propto\frac{1}{|i-m|}$, the farther away from the private clue word, the closer the temperature is to the constant temperature of KD.

\section{Privacy Clue Word Dictionary $\mathbb{P}$}\label{app:privacy_clue_word}

The construction of the privacy clue word dictionary $\mathbb{P}$ is very flexible and is not limited to private information in the traditional sense. For example, when distillation occurs across two institutions, privacy words may be related to project plans, progress, ideas, etc.
In addition, the size of $\mathbb{P}$ has a certain influence on the effect of distillation. If the size of $\mathbb{P}$ is too large, it may affect the effect of distillation. The privacy clue word does not necessarily have to be a single word, and can also be some special symbols, such as @ in an email address.

In our experiments, we construct a dictionary $\mathbb{P}$ of size 127. It mainly covers private information in terms of personal information, manually extracted from a subset of the Enron email dataset.

Examples of the privacy clue words in our constructed $\mathbb{P}$ are given in Table~\ref{tab:pcw}.
\begin{table*}[t]
    \centering
    \begin{tabular}{ccc}
    \hline
        Category & Privacy Clue Word & Example\\\hline
        Personal information & ID, password, name, birthday & The ID of Jeff is 23*****.\\
        Contact information & call ... at, @, address, fax & Please call me at 1-800-369.\\
        Location information & locate, business trip, find ... at & Allen is on a business trip to New York.\\
        Network information & https:/http:, IP, MAC & To change password, please go to: http://www.***.\\
    \hline
    \end{tabular}
    \caption{Examples of the privacy clue word dictionary $\mathbb{P}$.}
    \label{tab:pcw}
\end{table*}
\section*{Limitations}
Although swing distillation has significantly improved privacy protection, compared with fine-tuning, there is still room for performance improvement. We hence would like to investigate efficient ways of knowledge transfer from the teacher model to the student model in SD.

\end{document}